# PRISM-0: A Predicate-Rich Scene Graph Generation Framework for Zero-Shot Open-Vocabulary Tasks

Abdelrahman Elskhawy[1,2*†], Mengze Li[1,2*†], Nassir Navab[1] and Benjamin Busam[1]

[1*]Technical University of Munich (TUM), Munich, Germany.
[2]Carl Zeiss Meditec AG, Munich, Germany.

*Corresponding author(s). E-mail(s): a.elskhawy@tum.de; mengzelibitkth@gmail.com;
Contributing authors: nassir.navab@tum.de; b.busam@tum.de;
†Equal Contribution

**Abstract**

In Scene Graph Generation (SGG), structured representations are extracted from visual inputs as object nodes and connecting predicates, enabling image-based reasoning for diverse downstream tasks. While fully supervised SGG has improved steadily, it suffers from training bias due to limited curated data and long-tail predicate distributions, leading to poor predicate diversity and degraded downstream performance. We present PRISM-0, a zero-shot open-vocabulary SGG framework that leverages foundation models in a bottom-up pipeline to capture a broad spectrum of predicates. Detected object pairs are filtered, described via a Vision-Language Model (VLM), and processed by a Large Language Model (LLM) to generate fine- and coarse-grained predicates, which are then validated by a Visual Question Answering (VQA) model. PRISM-0's modular, dataset-independent design enriches existing SGG datasets such as Visual Genome and produces diverse, unbiased graphs. While operating entirely in a zero-shot setting, PRISM-0 achieves performance on par with state-of-the-art weakly-supervised models on SGG benchmarks and even state-of-the-art supervised methods in tasks such as Sentence-to-Graph Retrieval.

## 1 Introduction

Scene Graph (SG) represents visual relations in images as graphs, where nodes denote objects and edges represent pairwise relations, enabling rich visual understanding for tasks such as image captioning Zhong et al. (2020); Yang et al. (2023) and VQA Lei et al. (2023). Large-scale datasets like Visual Genome (VG) Krishna et al. (2017) have driven progress in supervised SGG, providing extensive annotated images and serving as the de facto benchmark. However, VG suffers from annotation bias: the top 50 predicates dominate most samples, while specific predicates are severely underrepresented Yu et al. (2023). This long-tail distribution, coupled with sparse graphs, under-represents the semantic richness needed for downstream reasoning.

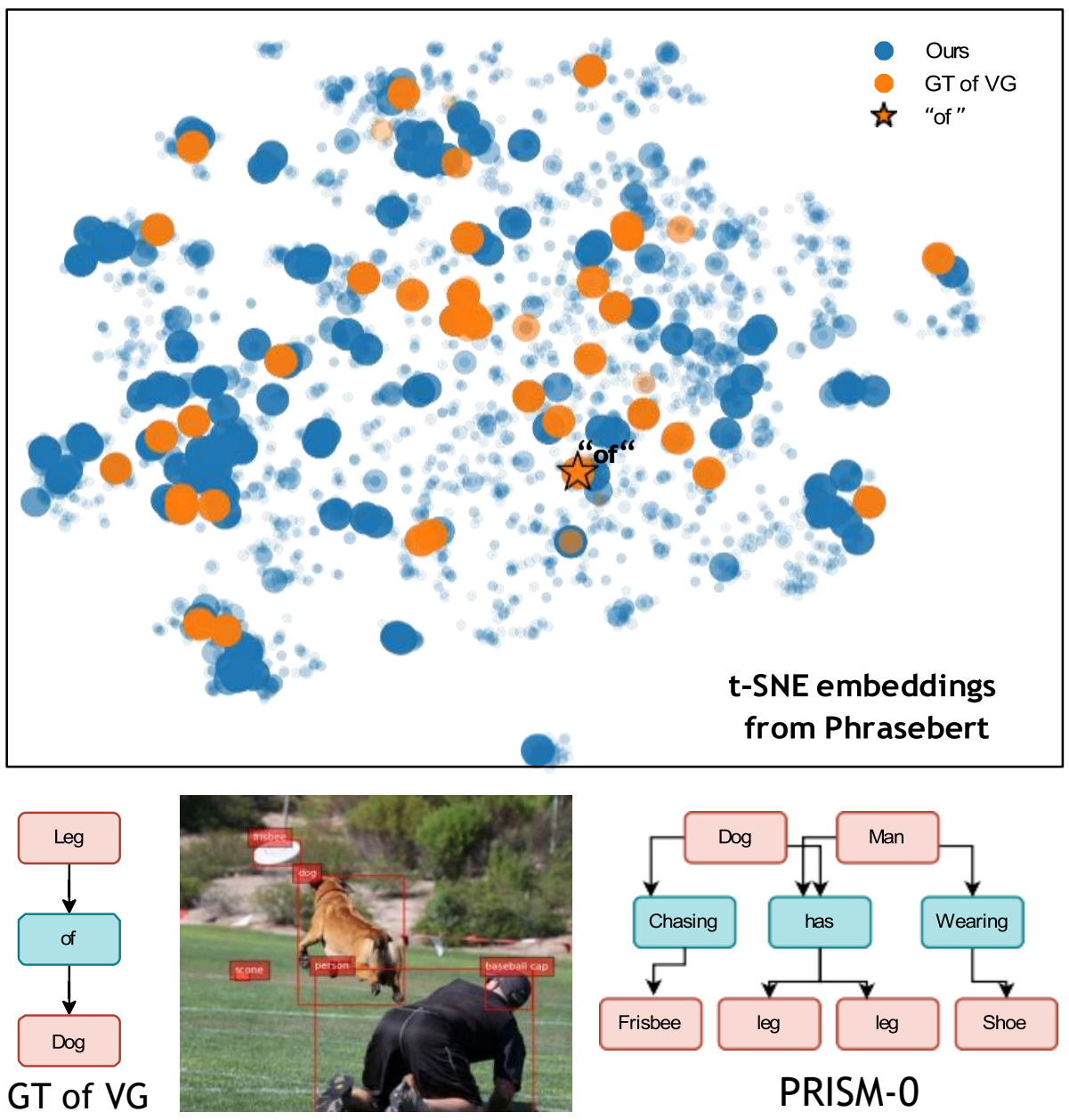

**Fig. 1 Predicate Distribution for VG Labels vs. PRISM-0.** ***Top:*** Predicate embedding visualization shows VG labels (orange) clustered around few predicates (e.g., "of", depicted as ∧), while PRISM-0 predicates (blue) populate the space more uniformly. ***Bottom:*** In an example image (middle), VG labels (left) are factually correct but miss semantic context, whereas PRISM-0 (right) captures both semantic and factual scene details.

As shown in Fig. 1 (top), our analysis of 7k VG images using PhraseBERT Wang et al. (2021) embeddings reveals that, despite accurate object detection, GT scene graphs in VG often contain trivial relations (e.g., *"leg of dog"*) that overlook higher-level semantics such as *"dog chasing frisbee"* or *"man playing with dog"*. Training on such annotations leads supervised SGG models to produce factually correct but semantically uninformative graphs during inference.

In this paper, we present **PRISM-0**, an efficient, modular, and scalable zero-shot open-vocabulary scene graph generation framework (Fig. 2), that leverages pre-trained VLMs and LLMs. While VLMs excel at captioning, they often lack the nuanced linguistic details needed for fine-grained predicate categorization. LLMs, on the other hand, encode rich relational knowledge from large-scale text corpora. PRISM-0 combines these strengths by prompting a VLM to generate descriptive captions for detected object pairs and parsing them with an LLM to produce both fine- and coarse-grained predicates, capturing a broad predicate spectrum through a bottom-up approach.

Unlike prior top-down methods Yu et al. (2023); Li et al. (2024) that depend on captions or ground-truth annotations, PRISM-0 directly constructs scene graphs from images, mitigating label bias and enabling novel, semantically meaningful predicates. To capture relationships missed in full-image captioning, we segment images into patches of varying sizes, allowing the VLM to focus on specific regions and extract more detailed relations. Our approach considers all object pairs, even those without overlapping bounding boxes, to capture critical non-overlapping interactions (e.g., *"person talking to another person"*, *"person watching TV"*) that define scene context.

To control computational cost in dense scenes, PRISM-0 employs filtering mechanisms to remove irrelevant or implausible relations. The **Geometric Filter** combines depth information with bounding boxes to estimate 3D object centers and discards spatially distant pairs. A final **Validation Module** further refines triplets based on semantic and spatial relevance in the image. We exemplify some PRISM-0 predictions together with the reference image and the VG annotation in Fig. 3.

Our main contributions are:

- We introduce PRISM-0, a modular, zero-shot open-vocabulary SGG framework that integrates VLMs and LLMs to generate fine- and coarse-grained predicates, benefiting from continual advances in foundation models.
- A bottom-up design that considers all object pairs including non-overlapping ones, combined with semantic and geometric filters to remove semantically or spatially implausible relations.
- A strategy to mitigate long-tail predicate bias and enhance label diversity by leveraging VLM captioning capabilities and LLM relational knowledge, with controllable predicate granularity.
- A novel diversity metric for relation predictions and extensive experiments on benchmarks, downstream tasks, and user studies demonstrating PRISM-0's ability to produce diverse, unbiased, and semantically meaningful scene graphs.

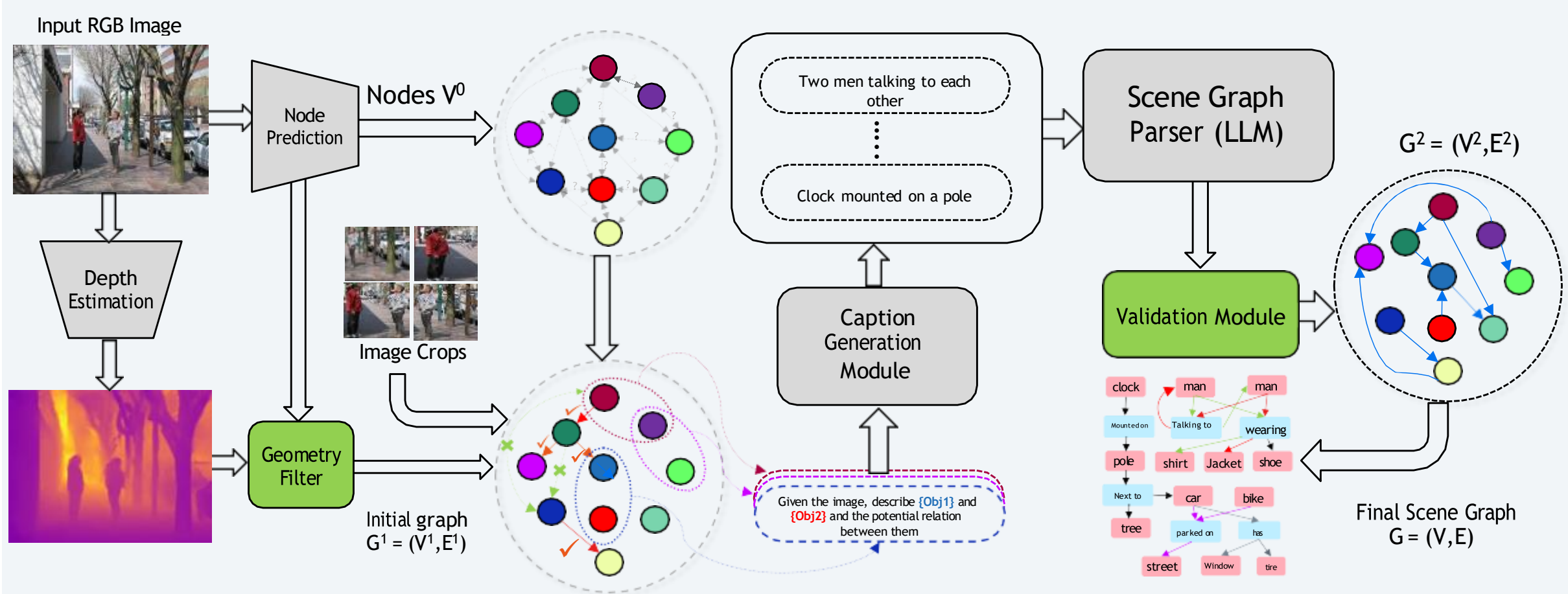

**Fig. 2** **Overview of the PRISM-0 framework.** The pipeline begins with node prediction and extraction of scene geometry from the image (top left). Pairwise captions are generated using a VLM and scene graph parsing is done with an LLM. The proposed geometry refines the relational pairs before triplet extraction. Finally, an uncertainty-based validation module evaluates the semantics and spatial relationship of the predictions, and provide ranking for relations to prioritize most plausible predictions.

# 2 Related Work

**Zero-Shot SGG:** A common way to bypass the bias in dataset annotations is to leverage foundation models at inference time. ELEGANT Zhao and Xu (2023) uses an LLM to propose relations for salient object pairs, while ConceptGraphs Gu et al. (2023) combines geometric cues with a VLM to produce open-vocabulary 3D graphs. These require no training but often emphasize spatial predicates and assume overlapping bounding boxes. CLIP-based visual relation detection methods Li et al. (2023); Chen et al. (2024) can predict relations in a zero-shot manner but do not perform a full SGG pipeline. CAPSGG Huang et al. (2025) introduces a capsule-based framework for zero-shot SGG, but unlike our approach, it is trained on the dataset and generalizes to novel subject-predicate-object combinations within a closed predicate vocabulary.

**Weakly-Supervised SGG:** Caption-guided approaches replace dense triplet labels with pseudo-relations parsed from datasets like MS-COCO or VG. CaCao Yu et al. (2023) prompts a language model to refine these captions, while LLM4SGG Kim et al. (2024) uses GPT-4 to extract triplets for detector training. PGSG Li et al. (2024) fine-tunes a vision–language model to generate textual graphs later parsed into triplets. Although these methods reduce annotation costs, their pseudo-labels inherit long-tail bias and limited predicate diversity from the source captions. In contrast, we adopt a bottom-up approach that directly obtains fine-grained captions from a VLM by iteratively focusing on each object pair. Some weakly supervised works Li et al. (2024); He et al. (2022) also report zero-shot results, but their setting differs by testing on both seen and unseen predicates from training.

**Unbiased SGG:** Fully supervised SGG inherits the long-tail bias of dataset annotations, motivating debiasing techniques. NICE Li et al. (2022) corrects noisy predicate tags before training, while DRM Li et al. (2024) uses a dual-granularity design to transfer knowledge from frequent to rare predicates, achieving state-of-the-art mean-recall. We include DRM as our main supervised baseline due to its predicate-rich output. However, even advanced debiasing methods remain constrained by annotation sparsity and long-tailed distribution eventually embedding dataset bias into the trained models.

# 3 Methodology

## 3.1 Problem Formulation and Method Overview

SGG aims to construct a structured graph $G = (V, E)$ representing an image $I \in \mathbb{R}^{H \times W}$, where nodes $V = \{v_i\}_{i=1}^{N}$ denote objects and directed edges $E = \{e_{ij}\} \subseteq V \times V$ encode pairwise relations. Each object $v_i$ has a label $o_i \in \mathcal{O}$ from the set of object classes $\mathcal{O}$ and a bounding box $b_i \in \mathbb{R}^4$ specifying its spatial extent. A relation $e_{ij} = (v_i, v_j, r_{ij})$ links objects $v_i$ and $v_j$ via a predicate $r_{ij} \in \mathcal{R}$ from the set of relation classes $\mathcal{R}$. The task is to jointly predict object labels and relations to capture both semantic and spatial interactions, producing a set of triplets $T = \{t_k\}_{k=1}^{K}$ that form the final scene graph.

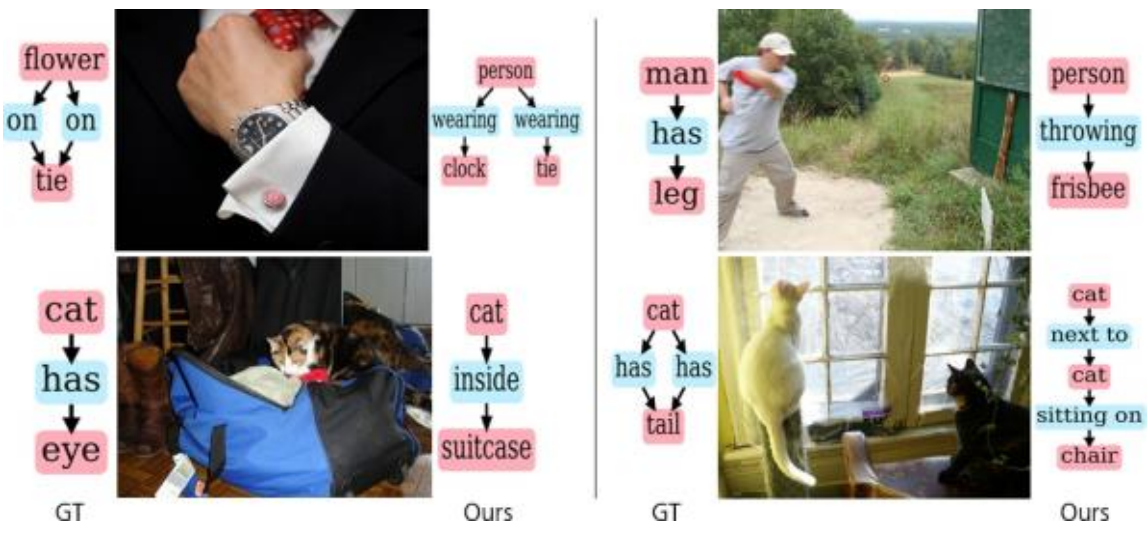

**Fig. 3 Qualitative Examples of Scene Graphs**: Input image (middle), Visual Genome's GT (left), and our predicted graph (right).

Our proposed framework (Fig. 2) follows a bottom-up paradigm to generate $G$ directly from $I$, without relying on captions or other ground-truth annotations. It comprises three main components: (1) node prediction, (2) caption generation, and (3) triplet extraction, plus two auxiliary modules: a geometry filter (GF) and a relation validation stage.

The pipeline begins with object detection, producing nodes $v_i$ with bounding boxes $b_i$ and labels $o_i$. For each object pair $(v_i, v_j)$, cropped regions $I_{ij}$ and prompts $p_i$ are processed by the captioning module $\mathcal{C}$ to generate relationship descriptions $c_i$, which an LLM parses into candidate triplets $t_i$. To reduce computation, the geometry filter GF combines 2D bounding boxes with median depth estimates to compute 3D centers and removes pairs exceeding a spatial threshold. A final relation validation stage uses a VLM to check each $t_i$ with the image, retaining only the most plausible and contextually relevant relations.

## 3.2 Node Prediction Module

The node prediction module detects objects in an input image $I$, providing both spatial and categorical information. It outputs a set of nodes $V = \{v_i\}_{i=1}^{N}$, each associated with a bounding box $b_i \in \mathbb{R}^4$ and an object label $o_i \in \mathcal{O}$, forming the basis for subsequent relation inference in the SGG pipeline.

## 3.3 Depth Estimation and Geometry Filter

Traditional SGG pipelines often select candidate object pairs based on overlapping bounding boxes, but this overlooks key non-overlapping interactions (e.g., *"Man, talking to, Man"* in Fig. 2). Enumerating all possible pairs would capture such cases but is computationally prohibitive in dense scenes, especially in our framework, which integrates multiple foundation models.

To address this, we introduce a depth estimation module and a geometry filter (GF) that leverage both 2D spatial layout and estimated 3D proximity. Given an input image $I$ and its depth map $D$, GF computes the 3D center of each bounding box using its 2D coordinates and median depth. For each object pair $(o_i, o_j)$, the relative 2D and 3D distances are combined to determine spatial relevance. A pair is retained only if it satisfies:

$$\lambda_1 \underbrace{(x/y)}_{\text{2D component}} + \lambda_2 \underbrace{\|d_i - d_j\|_2}_{\text{3D component}} < \tau, \qquad (1)$$

where $x$ is the distance between 2D bounding box centers, $y$ is the image diagonal length, and $d_i, d_j$ are normalized median depths for $o_i$ and $o_j$. The weights $\lambda_1, \lambda_2$ balance the contributions of 2D and 3D components, and $\tau > 0$ is the spatial relevance threshold.

By filtering pairs using both planar distance and depth-based proximity, GF removes spatially implausible relations before captioning, reducing computational load while preserving relational accuracy. This ensures the framework focuses on spatially coherent, contextually relevant interactions, improving efficiency without sacrificing important scene semantics.

## 3.4 Caption Generation Module

The Caption Generation Module $\mathcal{C}$ produces fine-grained relational descriptions for detected object pairs. Given $m$ detected objects $V =$ $\{v_1, v_2, \ldots, v_m\}$, we iterate over all unique pairs $(v_i, v_j)$.

While VLMs such as Up-Down Anderson et al. (2018) and BLIP Li et al. (2022) excel at holistic image captioning, they often overlook localized object-object interactions crucial for SGG. To direct attention to a specific pair, we generate localized image crops containing both objects and highlight them with distinct bounding box colors (red for $obj_1$, yellow for $obj_2$), following Shtedritski et al. (2023). We further guide the VLM using a structured language prompt. The VLM then processes each annotated crop, producing captions rich in semantic detail and relational context. These captions, containing both explicit and implicit relational cues, are passed to the Triplet Extraction Module for predicate identification. By visually cueing object pairs and linguistically steering the VLM, $C$ ensures relational descriptions are contextually precise and well-aligned with the objects' spatial and semantic relationships.

## 3.5 Triplet Extraction Module

Once object categories are detected, relational triplets are extracted efficiently using a LLM. We adopt a two-step prompting strategy to leverage the LLM's rich linguistic and contextual understanding.

Initially, the LLM paraphrases the pairwise captions to clarify ambiguous or imprecise expressions. For example, given a caption like *"two men engaged in a conversation"* (Fig. 2), directly extracting a triplet may produce vague outputs such as *"[two men, talking, each other]"*. Using Chain of Thought (CoT) prompting Wei et al. (2022), the LLM reformulates this into a clearer description: *"a man is talking to a man"*. This explicit paraphrase then enables precise extraction of the triplet *"[man, talking to, man]"*.

This two-step process improves the accuracy and clarity of relational triplet extraction, ensuring the generated triplets faithfully represent core visual interactions.

## 3.6 Relation Validation Module

To ensure the extracted triplets are both semantically and spatially consistent with the image, we propose an uncertainty-driven ranking module for relation verification. Unlike generic image-text alignment metrics such as CLIP-Score Radford et al. (2021), which often struggle with nuanced relationships Zhai et al. (2023), this module directly queries a VLM using a VQA approach.

For each candidate relation triplet, we construct a binary query to the VLM regarding the presence of the relation in the image. The model outputs probabilities for the responses "yes" and "no," where the normalized "yes" probability is interpreted as the confidence that the relation exists, and the corresponding "no" probability reflects the likelihood of its absence.

Repeating this for all candidate triplets, we rank relations by their confidence scores. Retaining low-confidence relations yields a more comprehensive scene graph, while discarding them results in a more precise graph emphasizing dominant relations.

# 4 Experiments

## 4.1 Experiments Configuration

**Datasets.** For SGG evaluation, we perform Predicate Classification on the VG test split Krishna et al. (2017), and use the generations to conduct a user study for subjective quality assessment. For Sentence-to-Graph Retrieval, following Tang et al. (2020), we use the 51k-image overlap between Visual Genome and MS-COCO Caption Lin et al. (2014), split into 35k/1k/5k train/val/test, and evaluate on gallery sizes of 1k and 5k as in the original setup.

**Implementation Details / Foundation Models.** For node prediction, we use Florence-2 (large-ft) Xiao et al. (2024) with SDPA Vaswani (2017) attention and no post-processing. Depth estimation is performed using Depth-Anything-V2 (base) Yang et al. (2024) for its fast and accurate inference. The geometry filter parameters are set to $\lambda_1 = 0.5$, $\lambda_2 = 1$, and $\tau = 0.45$.

For caption generation, we adopt LLaVA-OneVision Li et al. (2024) with batch size 15, maximum output of 50 tokens, and beam size 1. Triplet extraction module leverages LLaMA 3.2 (3B) AI (2024), using batch size 16 and a 256-token limit to process large caption subsets efficiently. For DBSCAN clustering we use $eps_final = 0.241$ and $min_samples = 5$.

## 4.2 Qualitative Evaluation

We qualitatively evaluate PRISM-0 by comparing generated SGs against GT annotations. A user study with 75 participants from diverse backgrounds was conducted, where each participant assessed 40 randomly selected images with corresponding SGs. Participants were shown a reference image accompanied by two SGs, PRISM-0 predictions and VG GT. Evaluations covered five criteria: overall perception (*Overall Situation*), node accuracy (*Object Names*), scene completeness, relational precision (*Relation Correctness*), and relational recall (*if the main relations are captured by the SG*).

As shown in Fig. 4, PRISM-0 consistently outperforms VG annotations across all metrics, with notable gains in semantic scene relations (completeness, situation, and main relation capture).

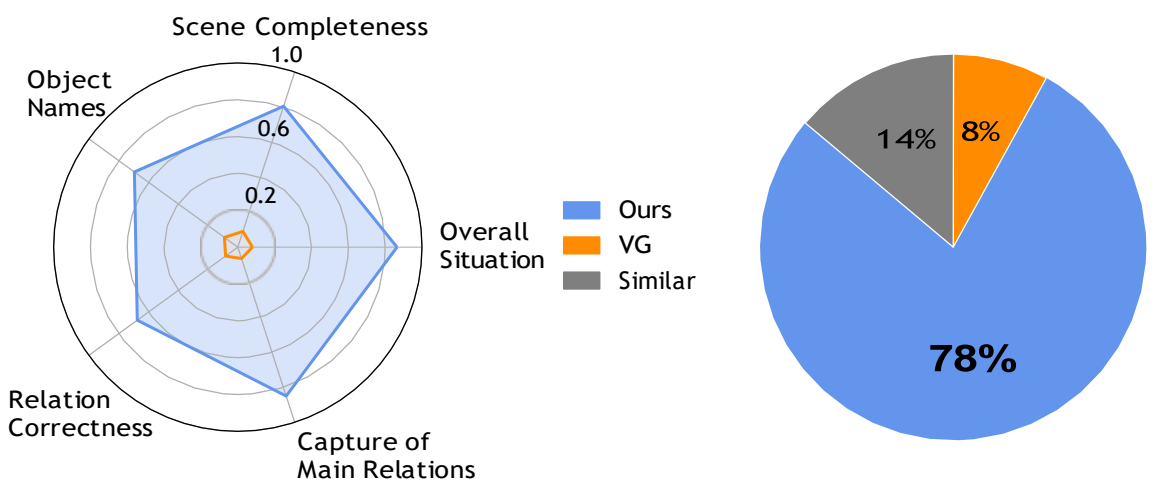

**Fig. 4 Subjective Quality of Generated SGs.** Left: Average normalized user ratings along five aspects for PRISM-0 predictions and VG annotation. Right: Preference distribution on overall quality of the SG; blue indicates PRISM-0 predictions are preferred, and orange indicates that VG annotations are preferred. The gray segment represents the percentage of users who found both methods equally descriptive.

**Statistical Analysis.** A paired sample t-test confirmed significant improvements in all five categories ($p < 0.05$), aligning with the qualitative feedback and underscoring the robustness of our approach.

## 4.3 Performance on Standard Benchmark: Predicate Classification (PredCls)

Beyond qualitative evaluation, we benchmark PRISM-0 on the standard VG PredCls task. To align open-vocabulary predictions with the closed-set evaluation, we use an LLM to map synonyms and reduce vocabulary mismatch. Moreover, we use the confidence-based validation module to rank triplets, to align the evaluation standard top@K recall. Unlike prior open-vocabulary methods Li et al. (2024); He et al. (2022) that train on base predicates and test on base + novel, PRISM-0 is fully open-vocabulary, treating all predicates as unseen, which complicates direct comparison since baselines benefit from supervised exposure to frequent predicates Li et al. (2024).

Table 1 reports Recall@K (R@K) and mean Recall@K (mR@K). While our Recall@50/100 slightly trails CaCao and PGSG, PRISM-0 surpasses all baselines on mR@50/100. This gap partly arises because PRISM-0 proposes more specific predicates (tail predicates) (e.g., *person-riding-bike* vs. VG's annotation *person-on-bike*), causing false negatives under closed-set evaluation metrics using biased annotations of VG as the ground-truth. Our goal is not to optimize closed-set benchmarks but to produce semantically richer, more diverse SGs. These results underscore how existing metrics undervalue fully open-vocabulary approaches.

**Table 1 Predicate Classification (PredCls) on VG.** *w*: weak supervision on VG base-predicate split, evaluated on base + novel. *zs*: zero-shot, no exposure to VG predicates.

| Model | Training | mR@50/100 | R@50/100 |
|---|---|---|---|
| CaCao | w | 10.3/12.6 | - |
| SVRP | w | 8.3/10.8 | **33.5/35.9** |
| PGSG | w | 10.8/13.9 | 26.9/33.9 |
| PRISM-0 | zs | **12.04/14.98** | 22.58/25.11 |

## 4.4 Diversity of Predicted Relations

Since standard metrics utilize a biased annotation set as the ground-truth, the unbiased predictions may be underrated. We therefore also evaluate the semantic diversity of the predicted relations. A robust zero-shot SGG method should produce a wide range of predicates and triplets rather than repeatedly predicting common ones. To measure this, we embed each predicted relation triplet into a vector space using the pre-trained language model SentenceBert Reimers and Gurevych (2019) and apply two diversity metrics:

(i) **Average Pairwise Distance (APD)**: the average cosine distance between all pairs of relation embeddings in the test set. Higher APD indicates greater semantic variety.

(ii) **DBSCAN Clustering**: we cluster predicate embeddings using DBSCAN and count the number of clusters. More clusters indicate broader coverage across distinct semantic relation groups. DBSCAN parameters are chosen to group very similar phrases together, estimating unique relation categories. We use *eps_final* = 0.241 and *min_samples* = 5.

**Table 2 Relation Diversity Analysis.** Higher APD and cluster count indicate broader embedding coverage; higher silhouette score reflects better cluster separation.

| Model | APD | DBSCAN | | |
|---|---|---|---|---|
| | | # clusters | Noise (%) | Silhouette |
| DRM w/o DKT | 0.76 | 52 | 23.3 | 0.038 |
| DRM | 0.78 | 75 | 30.6 | **0.231** |
| PGSG | 0.80 | 54 | 16.2 | 0.128 |
| PRISM-0 | 0.77 | 62 | 29.8 | **0.225** |

For a fair comparison, we also map our open-vocabulary relation predictions to the closed set before the diversity measurement.

Results in Table 2 compare PRISM-0 with PGSG Li et al. (2024), DRM Li et al. (2024), and DRM's raw outputs without debiasing (DKT). DRM without DKT shows the lowest semantic spread and fewest clusters, reflecting a bias towards head predicates. Its near-zero silhouette score indicates prediction collapse into a dense embedding region. Despite no predicate training or explicit debiasing, our zero-shot pipeline matches other unbiased methods, delivering diverse relation predictions, enabling predicate-rich zero-shot SGG.

## 4.5 How does dataset bias affect the results?

Our approach aims to mitigate bias toward frequent predicates that conventional supervised SGG models overfit to, yielding high recall on common predicates but poor performance on rare ones. To quantify this, we analyze the correlation between predicate frequency and per-class recall on the VG PredCls test set and show results in Table 3. For each predicate, we compute its frequency and the model's recall, then measure their dependency using distance correlation (dCor), where dCor = 0 indicates independence. Statistical significance is assessed via permutation testing to compute a p-value.

The correlation analysis reveals a stark contrast between our method and traditional SGG models. Fully supervised methods like DRM, even with debiasing, show strong correlation (dCor $> 0.5$, $p < 0.01$), indicating heavy dependence on predicate frequency and poor performance on rare predicates. Open-vocabulary methods that rely on VG supervision such as PGSG also exhibit moderate correlations. In contrast, PRISM-0 achieves a much lower dCor, demonstrating minimal bias from VG's long-tailed distribution. This suggests that a data-unbiased, fully open-vocabulary approach can inherently avoid frequency bias, unlike prior complex debiasing techniques.

**Table 3** Correlation: per-class recall vs. frequency

| Model | dCor | p-value |
|---|---|---|
| DRM | 0.65 | 0.001 |
| PGSG | 0.47 | 0.007 |
| PRISM-0 | **0.18** | **0.95** |

## 4.6 Sentence-to-Graph Retrieval (S2GR)

We validate PRISM-0's effectiveness by applying it to the downstream tasks of Sentence-to-Graph Retrieval (S2GR). Traditional SGG benchmarks such as PredCls are unsuitable for evaluating our zero-shot framework, as we explicitly avoid relying on Visual Genome (VG) annotations to overcome their bias and limited predicate diversity.

S2GR Tang et al. (2020) evaluates graph-level coherence by parsing human captions into text graphs, which serve as queries to retrieve scene graphs (SGs) representing images with similar visual concepts. This method uses only SGs as image representations, ignoring visual features. The query text-SGs are parsed from MS-COCO captions Lin et al. (2014).

Following Tang et al. (2020), we report mean-Recall@20/100 (R@20/100) and median rank

**Table 4** Performance comparison of Sentence-to-Graph Retrieval. ZS refers to the approach being fully zero-shot.

| | | **1000** | | | **5000** | | |
|---|---|---|---|---|---|---|---|
| **Model** | **ZS** | **R@20** | **R@100** | **Med** | **R@20** | **R@100** | **Med** |
| MTDE | C | 17.0 | 53.6 | 91 | 5.2 | 18.9 | 425 |
| PGSG | C | 48.4 | 75.3 | 97 | 27.1 | 54.3 | 313 |
| CaCao | C | 52.0 | 77.3 | 85 | 33.4 | 60.9 | 322 |
| LLM4SGG | C | **56.9** | **82.9** | **64** | **34.0** | **64.8** | **211** |
| **Ours** | ✓ | **56.9** | **80.2** | **71** | **32.6** | **64.0** | **223** |

(Med) on gallery sizes of 1K and 5K. We compare PRISM-0 against the Motif-based TDE (MTDE) Tang et al. (2020) and the recent supervised SGG methods: CaCao Yu et al. (2023), LLM4SGG Kim et al. (2024), and PGSG Li et al. (2024).

Table 4 shows that PRISM-0 outperforms MTDE, which struggles with predicate diversity, as well as PGSG and CaCao despite their training on MS-COCO captions. PRISM-0 ranks second to LLM4SGG which utilizes MS-COCO GT captions to generate weak labels, which are later used to train a fully supervised SGG model. This process results in their output being closely aligned with queries originally parsed from GT captions, yielding a higher retrieval accuracy.

## 4.7 Ablation Study

We assess the impact of key framework components on scene graph quality via Sentence-to-Graph Retrieval (S2GR), and report results in Table 5.

**Table 5** Ablation study for framework components. Evaluated on downstream task S2GR with a gallery size of 1000 images.

| Model Change | R@20 | R@100 | Med |
|---|---|---|---|
| w/o Geometric Filter | 29.8 | 61.1 | 239 |
| LLM (3B $\longrightarrow$ 1B) | 50.8 | 72.4 | 126 |
| Ours (Full Pipeline) | **56.9** | **80.2** | **71** |

**Geometric Filter.** Fig. 5 demonstrates how the geometric filter prunes irrelevant edges, retaining only meaningful relations. This improves downstream retrieval performance by reducing clutter and inaccurate relations. Removing this filter significantly degrades S2GR results (Table 5) and increases computational load on the VLM and LLM, slowing inference (Table 6).

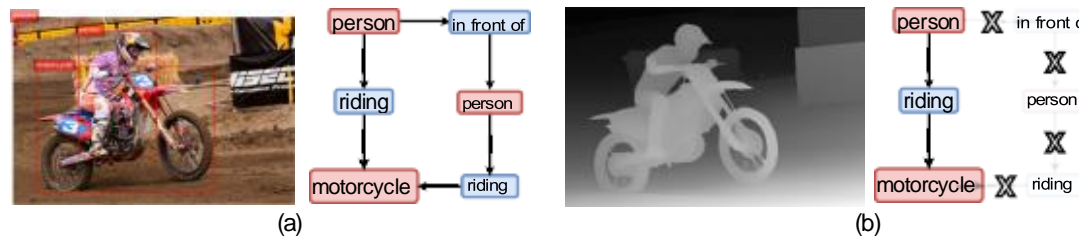

**Fig. 5** **Applying Geometric Filter** (a) shows an RGB image with detected objects including one person in the background (top left) and the predicted initial scene graph, including some trivial relations such as 'person in front of person'. (b) is the corresponding depth map with the filtered scene graph, where the relation between two people has been removed due to distance.

**Large Language Model.** Replacing the 3B llama 3.2 model with its 1B variant reduces performance (Table 5), indicating that LLM size impacts relation extraction accuracy and paraphrasing quality.

**Table 6** Ablation on per image inference time.

| Model | Avg. time (sec) |
|---|---|
| w/o GF | 90.0 |
| w/ GF | 67.5 |

**VG Bounding Boxes.** Using VG's GT bounding boxes instead of our node prediction module yields comparable performance, confirming that our detection results are on par with the GT annotations.

## 5 Limitations & Conclusion

While PRISM-0 demonstrates strong zero-shot performance in generating rich and comprehensive scene graphs, its effectiveness depends heavily on the accuracy of the node prediction module, making it vulnerable when object detection fails. Moreover, despite the use of filters such as the *Geometric Filter* and *Validation Module*, complex or ambiguous object relationships remain challenging. The bottom-up paradigm and the reliance on large-scale pre-trained models also introduces computational overhead, limiting applicability in real-time or resource-constrained settings. Nonetheless, PRISM-0's modular design, leveraging state-of-the-art Vision-Language and Large Language Models, addresses limitations of existing SGG datasets by mitigating sparse annotations avoiding dataset bias. PRISM-0 achieves improved performance on downstream tasks demanding high relational diversity, even in a fully zero-shot setting.

**Data Availability.** The work uses the open-sourced dataset Visual Genome, available at: https://homes.cs.washington.edu/ ranjay/visualgenome/index.html.

**Code Availability.** Not applicable

## Declarations

- **Funding** Not applicable
- **Competing interests** The authors have no competing interests to declare that are relevant to the content of this article.
- **Consent for publication** Not applicable

If any of the sections are not relevant to your manuscript, please include the heading and write 'Not applicable' for that section.